\documentclass{article}

\usepackage{arxiv}

\usepackage[utf8]{inputenc} %
\usepackage[T1]{fontenc}    %
\usepackage{hyperref}       %
\usepackage{url}            %
\usepackage{booktabs}       %
\usepackage{amsfonts}       %
\usepackage{nicefrac}       %
\usepackage{microtype}      %
\usepackage{lipsum}		%
\usepackage{graphicx}
\usepackage{natbib}
\usepackage{doi}

\usepackage{color,soul} %
\usepackage{arydshln} %
\usepackage{bbding} %
\usepackage{amssymb} 
\usepackage{amsmath}
\usepackage{mathrsfs} 
\usepackage{booktabs}

\title{Inertial Hallucinations -- When Wearable Inertial Devices Start Seeing Things}

\author{Alessandro Masullo, Toby Perrett, Tilo Burghardt, Ian Craddock, Dima Damen and Majid Mirmehdi\\
Department of Computer Science\\
University of Bristol\\
BS8 1UB, Bristol, UK\\
Email: a.masullo@bristol.ac.uk}

\renewcommand{\shorttitle}{\textit{arXiv} Template}

\hypersetup{
pdftitle={Inertial Hallucinations -- When Wearable Inertial Devices Start Seeing Things},
pdfsubject={},
pdfauthor={Alessandro Masullo, Toby Perrett, Tilo Burghardt, Ian Craddock, Dima Damen and Majid Mirmehdi},
pdfkeywords={Ambient Assisted Living, Multi-Sensory Fusion, Privileged Information, Modality Hallucination},
}

\begin{document}
\maketitle

\begin{abstract}
We propose a novel approach to multimodal sensor fusion for Ambient Assisted Living (AAL) which takes advantage of learning using privileged information (LUPI). We address two major shortcomings of standard multimodal approaches, limited area coverage and reduced reliability.  Our new framework fuses the concept of modality hallucination with triplet learning to train a model with different modalities to handle  missing sensors at inference time. We evaluate the proposed model on inertial data from a wearable accelerometer device, using RGB videos and skeletons as privileged modalities, and show an improvement of accuracy of an average 6.6\% on the UTD-MHAD dataset and an average 5.5\% on the Berkeley MHAD dataset, reaching a new state-of-the-art for inertial-only classification accuracy 
on these datasets. We validate our framework through several ablation studies.
\end{abstract}


\section{Introduction}
{A}{mbient} Assisted Living (AAL) is a novel technology that is changing the paradigm of healthcare from a clinic-based deployment to a home-monitoring setting. This shift in  healthcare is driven by the increasing trend of an ageing population living longer lives and chronic diseases consequently associated with it, like dementia, osteoarthritis, Parkinson's Disease and diabetes \cite{Ganesan2019}. The use of a combination of ambient sensors (e.g. cameras, depth sensors) and wearable sensors (e.g. accelerometers, gyroscopes, heart-rate monitors) deployed in patients' homes can help monitor their condition and assist clinicians to take important decisions without patients having to leave their home.

The advantages of multi-sensory fusion for AAL have been extensively addressed in the past, with different combinations of modalities leading to higher performance and reliability in many different tasks, for example action classification \cite{Tao2015, Majumder2020}, energy expenditure estimation \cite{Kalantarian2013, Calorinet2018}, and person re-identification \cite{Cabrera-Quiros2018, Masullo2020}. Such methods have an inherent requirement for the presence of all the modalities at both training and inference time and collecting multiple sensor modalities during a controlled experiment can provide a large datapool of rich information.

In a real AAL scenario, 
a multi-sensor network may be cumbersome to deploy, their protocols and transmission for synchronous analysis and recording may be challenging to implement, and not least of all, failed sensors in subjects' homes may not be re-activated as quickly and efficiently as in a controlled lab environment \cite{Woznowski2017}. 
This reduces the practical efficacy of in-the-wild multi-sensory systems, limiting their application to labs or rather intrusive, controlled home experiments. In addition, any individual sensor provides only partial coverage of the environment in which it is deployed, demanding the use of intelligent fusion strategies for the intersection of all the sensor ranges \cite{Masullo2021}.

\begin{figure*}
  \centering
  \includegraphics[trim=0 12cm 11cm 0, clip,width=0.75\linewidth]{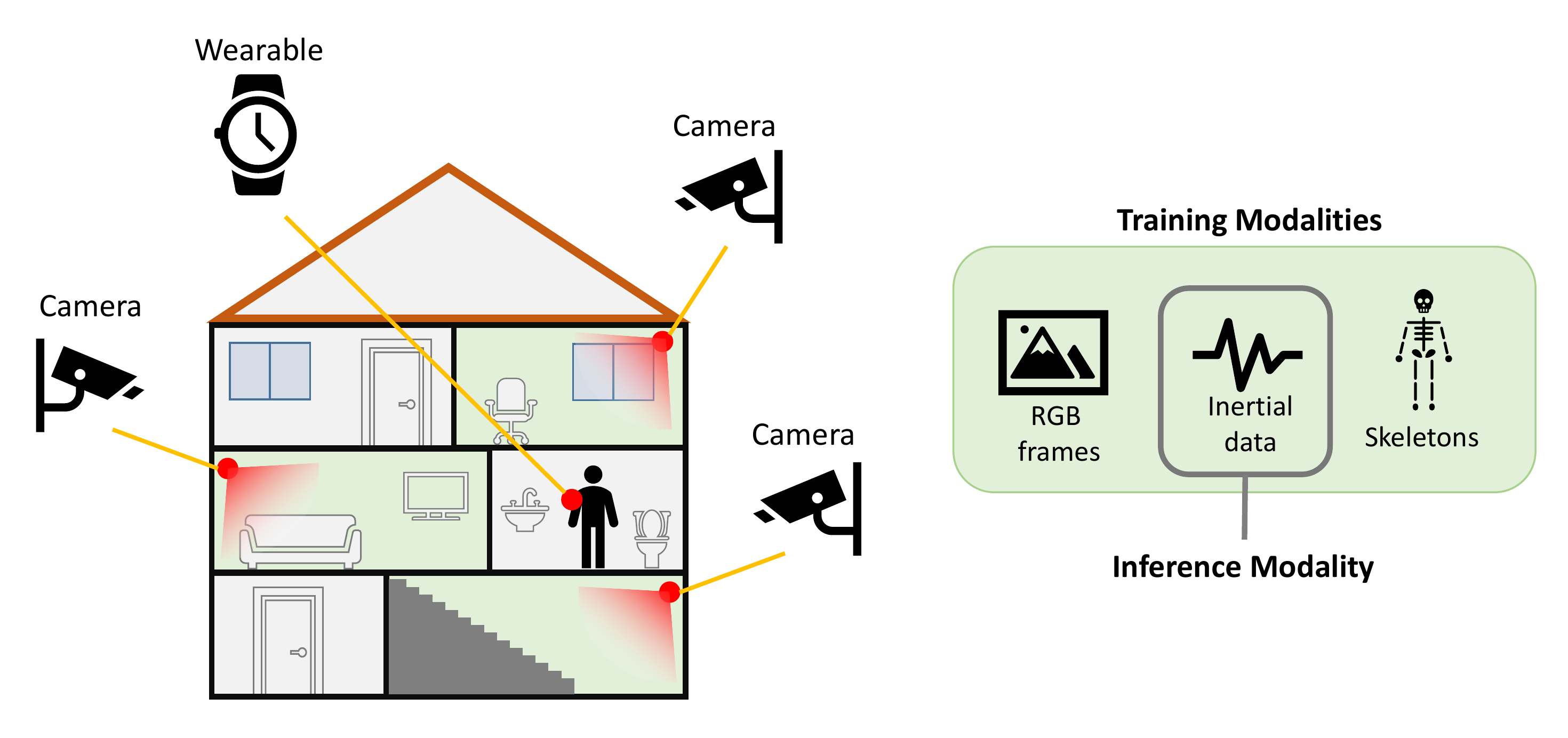}
  \caption{Typical distribution of sensors in an AAL framework: each camera is located in a different room and has a limited field of view, while the wearable is carried everywhere by the participant. Standard multi-modal approaches are only feasible in certain rooms (highlighted in green). With our proposed approach, multi-sensor data is used as privileged information when available and it is used to improve the accuracy of an inertial-only based model when other modalities are not available.} 
  \label{fig:description_method}
\end{figure*}

{The scenario presented in Figure~\ref{fig:description_method} exemplifies the AAL problem we wish to address}. {Wearable devices, producing inertial data, accompany  participants nearly everywhere but struggle to classify complex activities of daily living (ADL) \cite{Tao2015}. Cameras, on the other hand can boost ADL classification accuracy through sensory fusion \cite{Tao2015}, but only cover specific environments (either for privacy or practical restrictions). {This means that while inertial-based classifiers can be applied to every room in a house,} multi-sensory fusion algorithms are restricted to} rooms, highlighted in green in Figure~\ref{fig:description_method}, where subjects are in the range of view of cameras as well. We overcome this issue by replacing the missing cameras in certain environments with a \textit{hallucinated} version of it. 
{{{More specifically}, we address this by using solely the inertial data from a wearable device during inference, having employed multiple, fused modalities, e.g. inertial, alongside RGB frames and/or skeleton data from video as `privileged information' to train the model.} 
{This is achieved using a multi-stream hallucination approach based on a deep learning network 
that emulates features from different data streams to replace missing modalities during inference. }
At first, we train a standard multi-sensory network using all  available modalities, as depicted in Figure~\ref{fig:detailed_description_method} (left). 
Each stream is processed through a Convolutional Neural Network (CNN) backbone which generates encoded features for each modality. In a second stage (Figure~\ref{fig:detailed_description_method}, right), we create a hallucination network for each missing modality that takes as input the inertial data and hallucinates an output that are features of the missing modality. Our proposed method improves the state-of-the-art classification accuracy of an inertial-based classifier from 83.5\% to 90.1\% on the UTD-MHAD \cite{Chen2015a} dataset and from 86.0\% to 91.5\% on the Berkeley MHAD \cite{ofli2013berkeley} dataset. 

{Another key challenge for the deployment of AAL systems in homes is the computational effort. Modality hallucination requires the processing of a different hallucination network for each missing modality. The number of models required to run soars when multiple sensors are deployed in a house and can be incompatible with the computational power that is typically available in a real AAL house \cite{Woznowski2017}. To tackle this problem, we propose a novel and more efficient way of hallucinating missing modalities that emulates a feature fusion layer directly from the inertial data, rather than individual modalities.} Our paradigm is based on learning using privileged information (LUPI) \cite{Vapnik2009}. LUPI is commonly applied to problems where the source and the target domains are similar to each other \cite{Yosinski2014}. To the best of the authors' knowledge, our work constitutes the first time it has been applied to an AAL application and we propose a novel way of training our hallucination networks and generating hallucinated features.} %

{The contributions of this work are: (i) we propose a method to solve two of the biggest drawbacks of standard multi-sensory approaches for AAL systems, i.e. limited area coverage and reduced reliability; (ii) we present a novel training strategy using privileged information that takes advantage of triplets to achieve better hallucination performances when compared to standard regression; (iii) we show how accelerometer data trained with privileged information from visual data (i.e. RGB and skeleton data) significantly improves action classification performance; (iv) we propose a different approach to modality hallucination that sacrifices accuracy in exchange of better computational efficiency for AAL systems; (v) we show a new state-of-the-art in classification accuracy for inertial sensors on two different datasets: UTD-MHAD \cite{Chen2015a} and Berkeley MHAD \cite{ofli2013berkeley}.}

\section{Related work}
\label{sec:related_work}
In this section, we consider the most recent works on multi-sensor fusion for AAL applications, followed by  those that offer different strategies to handle missing modalities, for example by replacing lost modalities with synthetic data or using the concept of privileged information. %

{\bf Multi-sensory fusion for AAL }
Fusing multiple sensor modalities, particularly aimed at AAL applications, is already a mature approach, with some common combinations being RGB+inertial \cite{Tao2015, Martinez-Villasenor2019, Imran2020a, Islam2020}, RGB+depth \cite{Shahroudy2016, Vielzeuf2019, Islam2020}, depth+inertial \cite{Chen2016a, Dawar2018a, Ahmad2019}, and video silhouettes+inertial \cite{Calorinet2018, Masullo2020}. These are typically aimed at Human Action Recognition (HAR), fall detection, energy expenditure estimation and more. The majority of these works show a net improvement in their performance scores when multiple modalities are fused together; this is often due to different modalities exploiting a range of clues from human activities that help to better understand its complexity. One of the strongest limitations of standard multi-sensory fusion algorithms is that they require the presence of all modalities at inference time. %
For more details on multi-sensor fusion for HAR, including different data modalities, fusion strategies and datasets available, the survey from Aguileta et al. \cite{Aguileta2019} summaries many of the most important works in the field, while more recently, Majumder et al. presented a review on the more specific aspects of visual and inertial fusion for HAR \cite{Majumder2020}.

{\bf Handling a Missing Modality } The potential lack of a sensor modality in a multi-sensor network during inference can also be formulated as a missing modality problem. 
In \cite{Neverova2016}, this is dealt with by carefully fusing multiple streams through a gradual fusion that involves random dropping of separate channels. The method, dubbed \textit{ModDrop}, is able to produce robust classifiers that compensates for any number of missing modalities during inference. A similar approach is also followed by Hossain et al. \cite{Hossain2020}, where missing values in the training dataset are artificially induced to deal with potential missing data during inference, improving robustness and generalisation of the model. In \cite{Diethe2017}, Diethe et al. investigate %
heterogeneous sensor modality fusion in an AAL setting and propose a Bayesian model to explore the weight of each modality to enhance the interpretability of their fusion model. A different approach is followed by Saeed et al. \cite{Saeed2018}, who take advantage of adversarial autoencoders for handling missing modalities by replacing the missing sensors with synthetic data generated from the available modalities. Lee et al. \cite{Lee2020} use multimodal latent variables and a variational product-of-experts approach to automatically detect corrupted sensors and also replace them with synthetic data generated from the remaining sensors. {While the use of synthetic data to replace a missing modality can be viable for somewhat similar modalities like RGB and depth \cite{Hoffman2016}, it is arguably a much more difficult case when for example RGB frames need to be synthetised from inertial data, which would be the case in our AAL scenario. }%

{\bf Modality Hallucination } %
The paradigm of using privileged information was first presented by Vapnik et al. \cite{Vapnik2009}, who described its first application to Support Vector Machines (SVMs) and the theoretical framework behind it. %
Closer to the objectives of our work, Hoffman et al. \cite{Hoffman2016} proposed an extension of the LUPI framework to CNNs, introducing the concept of modality hallucination. In a first stage, they use both RGB and depth data to train a Fast R-CNN network \cite{Girshick2015} for object detection.
In their second stage, they freeze the depth stream of their network and train a hallucination network that takes RGB data for input and produces depth features for output. At inference time,  RGB only is fed through both the initial RGB stream and the depth-hallucination network. Even though this approach looks similar to the problem of depth estimation from RGB \cite{Hambarde2020}, Hoffman et al. demonstrated that hallucinating depth features is, instead, a much better way to use the depth data for object detection. More recently, Garcia et al. \cite{Garcia2020} tackled a similar problem where depth features were hallucinated through adversarial learning rather than regression. %

{In this paper we leverage the concepts of LUPI and modality hallucination to develop a novel framework that is suitable for AAL monitoring scenarios and improves over both previous applications of modality hallucination and alternative missing modality approaches. A broad summary of the most pertinent related works can be found in Table~\ref{tab:related_work}.}

\begin{table*}
 \caption{Summary of main related works.}
  \centering
\begin{tabular}{@{}lll@{}}
\toprule
\textbf{Reference}                & \textbf{Framework}                & \textbf{Method}                                             \\ \midrule
Moddrop \cite{Neverova2016}       & Missing Modality                  & Gradual fusion with random dropping                         \\
Hossain et al. \cite{Hossain2020} & Missing Modality                  & Artificial missing values                                   \\
Saeed et al. \cite{Saeed2018}     & Reconstructing Missing Data       & Adversarial autoencoders                                    \\
Lee et al. \cite{Lee2020}         & Reconstructing Missing Data       & Multimodal latent variables, variational product-of-experts \\
Hoffman et al. \cite{Hoffman2016} & Modality Hallucination            & Fast R-CNN                                                  \\
Garcia et al. \cite{Garcia2020}   & Modality Hallucination            & Adversarial learning                                        \\
Ours                              & Integrated Modality Hallucination & Triplet learning                                            \\ \bottomrule
\end{tabular}
  \label{tab:related_work}
\end{table*}

\section{Materials and methods}

\subsection{Modality Hallucination}
\label{sec:modality_hallucination}
Inspired by the idea of modality hallucination in \cite{Hoffman2016}, we develop a multi-stream architecture to handle missing modality management in sensor fusion architectures. {In this section, we explore the theoretical framework of modality hallucination. We present its standard form, hereafter referred to as \textit{Individual Hallucination}, and a novel strategy that we call \textit{Integrated Hallucination}.} {Although our theoretical framework  can be applied to any general number of modalities and combinations, in application this work will focus only on inertial data supplemented with RGB videos and skeletons.} 

\begin{figure*}
  \centering
  \includegraphics[trim=0 13cm 3cm 0, clip, width=0.9\linewidth]{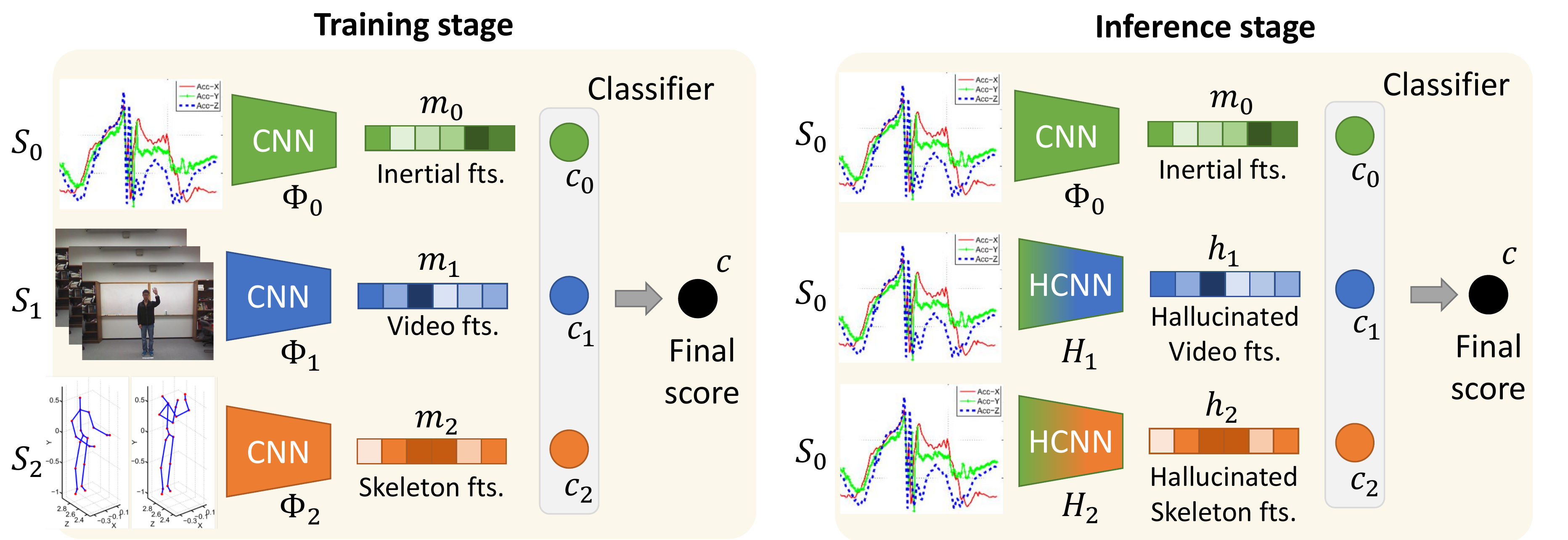}
  \caption{Individual Modality Hallucination for late fusion -- During training (left), all the data available (RGB, skeleton and inertial) is used to train an action classifier. The fusion of multiple streams is performed through a late fusion strategy. During inference (right), only inertial data is used to run the models. The inertial data is fed through the hallucination networks to generate hallucinated features which emulate the original features and are used to classify the actions.%
  }
  \label{fig:detailed_description_method}
\end{figure*}

\paragraph{Individual Hallucination} Our network comprises $N$ different streams $S_i, i\in\{0, 1, \dots, N\}$ %
(e.g. videos, accelerations, PIR activations, etc. 
as depicted in Figure \ref{fig:detailed_description_method}. %
For each stream $S_i$ we can build, {for simplicity}, an architecture $\Phi_i$ that solves  AAL-related downstream tasks %
 (e.g. action classification, or energy expenditure, etc.), with score $c_i$. {For a classification task, we have}
\begin{equation}
\label{eq:single_stream}
\begin{gathered}
m_i = \Phi_i(S_i\ ;\ \theta_{S_i}),  \\
c_i = \textrm{Softmax}\left(\textrm{FC}\left(m_i\right)\right),
\end{gathered}
\end{equation}
where $\theta_{S_i}$ are the weights of the model $\Phi_i$, a CNN backbone that produces a set of features $m_i$, and FC is a fully connected layer. Let the first stream $S_0$ be the \textit{inference modality} to indicate the only modality that will be available at inference time\footnote{The inference modality is used alone only when no other modality is available at inference time. If multiple modalities are available, then we will still use them all in a standard fusion approach.} and the remaining $S_i, i\in \{1 \dots N\}$, be \textit{privileged modalities}. In the first stage of training, all the $N$ models are trained individually, 
as shown in Figure~\ref{fig:detailed_description_method} (left). %
These $N$ models are then combined in a fusion model $\mathcal{F}$ that produces the final output $c$, 
\begin{equation}
\label{eq:fusion_network}
    c = \mathcal{F}\left(m_0, m_1, \dots, m_{N}\ ; c_0, c_1, \dots, c_N ;\ \lambda_\mathcal{F}\right).
\end{equation}
$\mathcal{F}$ %
can be as simple as a late fusion strategy (in which case $\lambda_F$ is not needed), as pictured in Figure~\ref{fig:detailed_description_method} (left). Alternatively, $\mathcal{F}$ can be a more complex mid-fusion strategy, 
where $\lambda_F$ constitutes an additional set of parameters %
that are needed to combine the $\Phi_i$ networks.  %
More details on the fusion strategies are presented in Section~\ref{sec:fusion_strategies}.

In the second stage of training, %
we freeze the weights of the models $\Phi_i$. %
Then, for each of the privileged modalities $S_i, i\neq0$, we build a hallucination model $H_i$ that takes as input the inference modality $S_0$ and produces as output {the hallucinated features $h_i$, which we train towards matching $m_i$. The hallucination models are still convolutional networks, but they produce features as if the input was a different modality, hence Hallucinated Convolutional Neural Network (HCNN).} %
Similarly to  Eq.~\ref{eq:single_stream}, %
\begin{equation}
\label{eq:hallucination_model}
    h_i = H_i\left(S_0\ ; \theta_{H_i}\right) \approx m_i .
\end{equation}
Once the parameters $\theta_{H_i}$ are trained, the hallucinated features $h_i$ can be used in conjunction with the inference modality $S_0$ using the fusion network $\mathcal{F}$ from Eq.~\ref{eq:fusion_network}, such that %
\begin{equation}
\label{eq:hybrid}
    c = \mathcal{F}\left(m_0, h_1, \dots, h_{N}\ ; c_0, c_1, \dots, c_N ;\ \lambda_\mathcal{F}\right) ,
\end{equation}
with the important difference that the only input required in Eq.~\ref{eq:hybrid} is the inference modality $S_0$. {The inference stage of Individual Modality Hallucination is depicted in Figure~\ref{fig:detailed_description_method} (right) for the implementation of a late fusion strategy.}

\begin{figure*}
  \centering
  \includegraphics[trim=0 13cm 0cm 0, clip, width=0.9\linewidth]{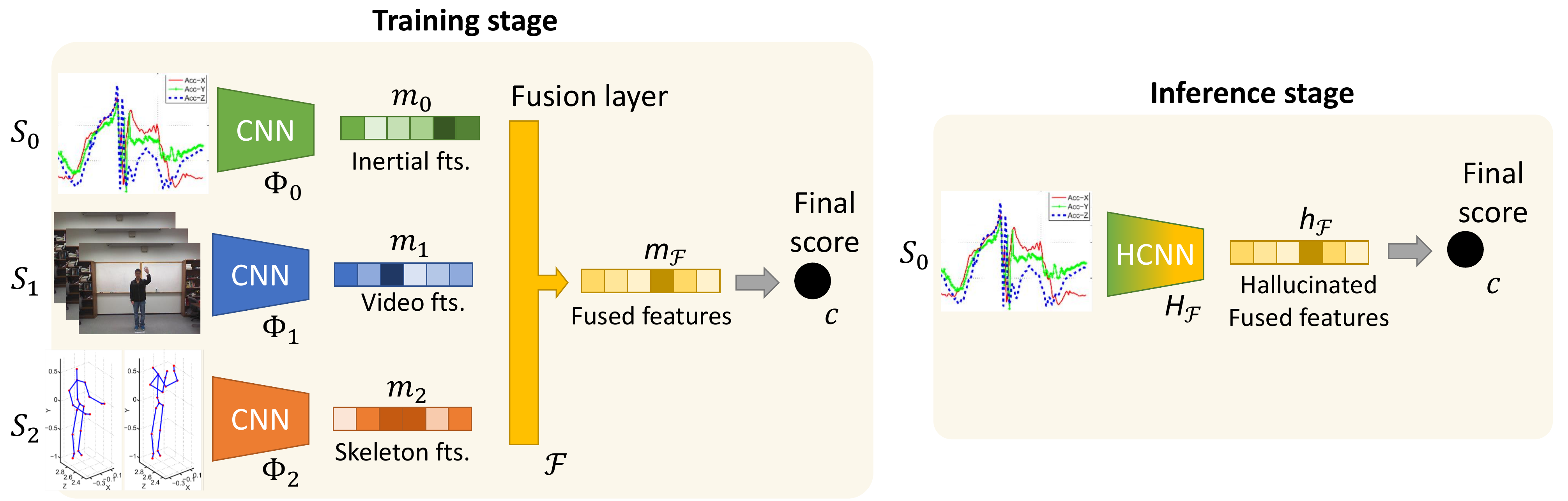}
  \caption{Integrated Modality Hallucination -- During training mode (left), all the data available (RGB, skeleton and inertial) is used to train an action classifier. The fusion of multiple streams is performed through a feature fusion strategy, which generates an additional set of fused features. During inference (right), only inertial data is used to run the model. The inertial data is fed through the hallucination networks to generate directly fused features, which are then used for the classification.}
  \label{fig:integrated_hallucination}
\end{figure*}

\paragraph{Integrated Hallucination} %
With the individual hallucination strategy, each of the privileged modalities $S_i$ requires its own hallucination model $H_i$ that needs to be trained independently. As we will see, training a hallucination model can be very time consuming and for a large number of sensor modalities this can be impractical. If we consider {the mid-fusion model from} Eq.~\ref{eq:fusion_network}, 
let %
$m_\mathcal{F}$ be the features produced by $\mathcal{F}$ through the fusion of the $m_i$ individual streams' features, as depicted in Figure~\ref{fig:integrated_hallucination} (left). %
{We can now create a single hallucination network $H_\mathcal{F}$, producing the features $h_\mathcal{F}$ in output, which we train towards matching $m_\mathcal{F}$ directly instead of the individual $m_i$:}
\begin{equation}
    h_\mathcal{F} = H_\mathcal{F}\left(S_0\ ;\ \theta_{H_\mathcal{F}}\right) \approx m_\mathcal{F} .
\end{equation}
{At inference time, the hallucination model $H_\mathcal{F}$ takes as input the  modality $S_0$ and produces as output the fused features $m_\mathcal{F}$ directly,} %
as shown in Figure~\ref{fig:integrated_hallucination} (right). In Section~\ref{sec:results} we will show that this strategy sacrifices  performance in exchange for ease of training and lower computational expense.

\subsection{Training Hallucination Networks}
\label{sec:training_hall_net}
\begin{figure*}[h]
  \centering
  \includegraphics[trim=1cm 17cm 1cm 0, clip,width=0.85\linewidth]{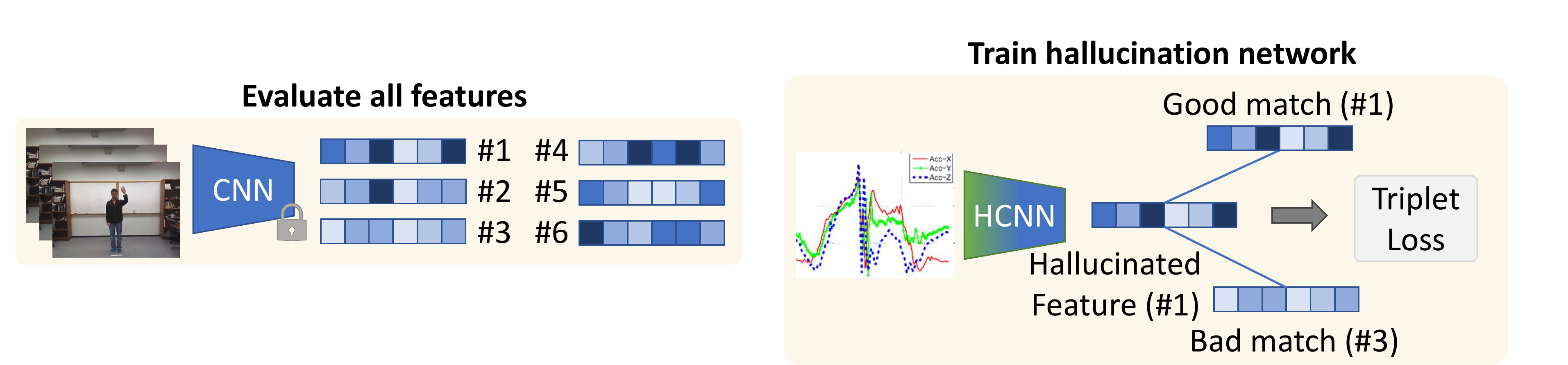}
  \caption{The hallucinated networks are trained using triplets. For the RGB hallucination network, illustrated in the figure, video features are pre-computed for the entire dataset (left). Inertial data is then used to simulate those features by showing pairs of good and bad matches that are then fed into the triplet loss (right).}
  \label{fig:triplet_description}
\end{figure*}

Once the single stream models $\Phi_i$ and the fusion model $\mathcal{F}$ are trained, the hallucination models $H_i$ can be learned. Since the model that we are trying to imitate, hereafter \textit{target model}, is frozen during the training of the hallucination networks, it is very convenient to pre-compute all the features $m_i$ for the entire dataset to speed up the training process (Figure~\ref{fig:triplet_description} left). The hallucination model training approach in \cite{Hoffman2016} %
{uses a loss function that combines the default losses for their numerous models (e.g. classification, bounding boxes, etc.) with an additional hallucination loss. They combine a total of 11 different losses, including 5 categorical cross-entropy losses, 5 smooth L1 losses and one additional hallucination loss, that are carefully balanced with  loss weights that must be manually optimised. Such optimisation can be very tricky and, in our experiments, we found it to be very inefficient. Thus, we propose a novel method to train our hallucination network that is not only easier to implement, but it also performs better than  \cite{Hoffman2016}'s regression loss.}

\paragraph{Triplet learning} %
Let us consider a sample $j$ from our dataset and the $j^{th}$ triplet $\mathcal{T}_j$, defined as:
\begin{equation}
    \mathcal{T}_j = \left(\text{anchor}, \text{positive}, \text{negative}\right) \equiv \left(h_j, m_j, m_k\right) ~,
\end{equation}
where $h_j$ are the features produced by $H$ from Eq.~\ref{eq:hallucination_model}, $m_j$ are the features of the target model for the same sample $j$, and $m_k$ are features from the same target model but for a different sample $k$, as shown in Figure~\ref{fig:triplet_description} (right). For this triplet, the Triplet Loss $\mathcal{L}_\mathcal{T}$ \cite{Schroff2015} %
is defined as %
\begin{equation}
    \mathcal{L}_\mathcal{T} = max\left\{\left|h_j-m_j\right|^2 - \left|h_j-m_k\right|^2 + a, 0\right\} ~,
\end{equation}
where $a=0.2$ is a constant. 

\paragraph{Skeleton proxy} Another important aspect that needs to be considered when training hallucination video features from inertial data is the choice of the target space. Features derived from video, optimised to perform an action classification task, will likely include both clues from the movement of the subject (e.g. posture, speed, dynamics, etc.) and visual clues from either the objects they are interacting with or the background. The same task performed using inertial sensors, will inherently produce features that only contain information about the motion of the subjects (or lack thereof). It is therefore potentially counterproductive to imitate the entire feature space produced by the video input. %
What if there was a sub-space of the video data that only includes information about the movement and posture of the subjects? Enter the skeleton. Skeleton data, including positional information for each joint of the human body, constitute the ideal proxy of video for the inertial features to imitate. In fact, the derivatives of the joint positions are related to the joint velocity and acceleration, which is exactly what inertial sensors are designed to measure.

\subsection{Fusion strategies}

\label{sec:fusion_strategies}
In order to find the optimal strategy for aggregating the available data modalities, we tested different combinations of fusion strategies. First, we considered a \textit{late fusion} approach where each data stream $\Phi_i$ is trained individually to produce the classification scores $c_i$, as per Eq.~\ref{eq:single_stream}. %
In a late fusion strategy approach, the final classification score $c$ is the product of the individual $c_i$ classification scores:
\begin{equation}
c = \prod_{i=1}^{N} c_i.
\end{equation}
We also investigated a \textit{mid-fusion} approach, where the intermediate features $m_i$ generated by each sensor modality are combined together prior to classification. The combination of the features can be performed in a variety of ways, with concatenation being a common choice:%
\begin{equation}
\begin{gathered}
m_F = \text{concatenate}\left(m_0, \dots, m_{N}\right) \\
c = \textrm{softmax}\left[\textrm{FC}\left(m_F\right)\right] ~,    
\end{gathered}
\end{equation}
where $m_F$ is an additional set of features that represents the fusion of the $m_i$ features.%

We also considered a further mid-fusion strategy where the concatenated features are fed into a fully connected layer before being processed by the classification layer:
\begin{equation}
m_F = \text{FC}\left[\text{concatenate}\left(m_1, \dots, m_{N}\right)\right].
\end{equation}
We will refer to this fusion strategy as \textit{mid-dense}. If the number of fully connected units for the $m_F$ vector is chosen to be the same as the number of elements of $m_i$, this fusion strategy has the advantage that it can be used as a direct target for hallucination, as already shown in Section~\ref{sec:modality_hallucination} when referring the the integrated hallucination. A summary of the different fusion strategies and combinations with hallucination modalities is available for convenience in Table~\ref{tab:fusion_hall_modes}.

\begin{table}
 \caption{Possible combinations of fusion strategies and hallucination modes}
  \centering
\begin{tabular}{rc|c}
\multicolumn{1}{l}{}                          & \multicolumn{2}{c}{\textbf{Hallucination Mode}} \\
\multicolumn{1}{r|}{\textbf{Fusion strategy}} & Individual Hallucination        & Integrated Hallucination        \\ \cline{1-3} 
\multicolumn{1}{r|}{Late}                     & \checkmark                         & N/A                    \\
\multicolumn{1}{r|}{Mid Concatenate}        & \checkmark                         & N/A                    \\
\multicolumn{1}{r|}{Mid Dense}              & \checkmark                         & \checkmark                  
\end{tabular}
  \label{tab:fusion_hall_modes}
\end{table}

\subsection{Data pre-processing and augmentation}
\label{sec:preprocessing}
\begin{figure*}%
\centering

\parbox{0.31\textwidth}{\includegraphics[width=\linewidth]{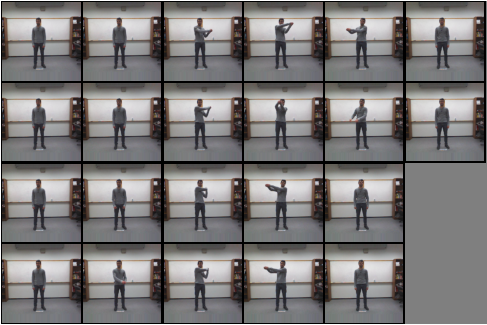}}%
\qquad
\begin{minipage}{0.30\textwidth}%
\includegraphics[width=\linewidth]{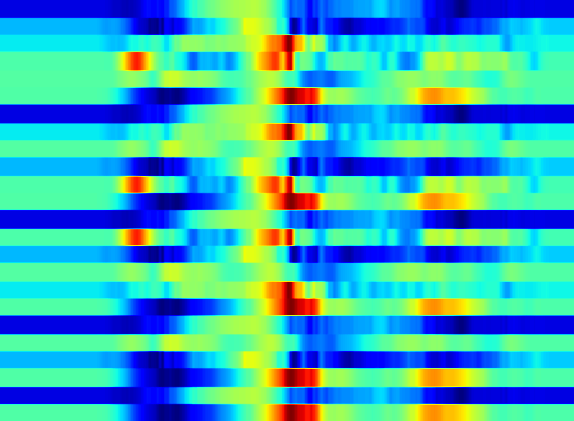}
\end{minipage}%
\qquad
\begin{minipage}{0.30\textwidth}%
\includegraphics[width=\linewidth]{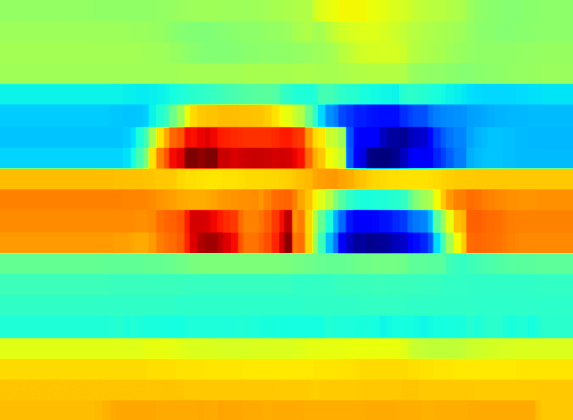}
\end{minipage}%

\caption{Example images of signals used as input for our networks. Images are generated from (left) RGB data (sub-sampled for clarity) (middle) inertial stream and (right) skeleton stream from the UTD-MHAD \cite{Chen2015a} dataset.}
\label{fig:inertial_image_sample}
\end{figure*}

The main streams used in this work are video, inertial and skeleton data: 

\paragraph{Video stream} The videos were randomly cut into shorter clips of 64 frames. If the input videos were shorter than 64 frames, additional frames were included by repeating the first and last frames with a random balance. Furthermore, videos were resized to 256x256 pixels while keeping the aspect ratio and randomly cropped into 224x224 patches for augmentation. A sample from the RGB videos in the dataset is presented in Figure~\ref{fig:inertial_image_sample}(a).

\paragraph{Inertial stream} Data from the inertial sensors include accelerometer and gyroscope measurements synchronised with the video stream. The inertial data was first cropped in a similar fashion to the video stream (where 64 frames are equivalent to 217 inertial samples) and then each sensor (i.e. gyroscope and accelerometer) was normalised between $\pm 1$. We followed  \cite{Jiang2015a}'s strategy to create an ``activity image'' from different inertial sensors %
by repeatedly stacking each of the three channels $(x, y, z)$ into a new arrangement where every signal has the chance to be adjacent to every other signal at least once. For 2 different sensors $S_1$ and $S_2$, with channels numbered as $S_1 = (1,2,3)$ and $S_2 = (4,5,6)$, we rearranged the data in a single element $\hat{S}$ %
arranged as:
\begin{equation}
    \hat{S} = \left(1,2,3,4,5,6,1,3,5,2,4,6,1,4,2,5,3,6,1,5,2,6,1,6\right).
\end{equation}
The resulting modality $\hat{S}$ was then reshaped into an image of 224x224 pixels as depicted in Figure~\ref{fig:inertial_image_sample}(b). %

\paragraph{Skeleton stream} Skeleton sequences were also cropped to 64 frames with shorter sequences being supplemented with additional copies of the first and last skeletons. Joint positions were then normalised between $\pm 1$, and finally rearranged next to each other to form an image that was resized to 224x224 pixels. Unlike the inertial data, skeleton joints were simply arranged according to the order in which they were produced by the skeleton extractor, as shown in Figure~\ref{fig:inertial_image_sample}(c). For example, for 32 joints, the possible adjacent sequences are more than 1000 and this information would be lost anyhow when the image is resized to be compatible as input to the network model. In practice, the skeleton image generated in this simple way produces excellent performances. We note that optimal skeleton pre-processing techniques are not within the scope of this work.

\subsection{Datasets}
We tested our algorithms on two different datasets, UTD-MHAD \cite{Chen2015a} and Berkeley MHAD \cite{ofli2013berkeley}.
{UTD-MHAD} was recorded to develop research for human action recognition using the fusion of multiple sensors from different modalities. It includes RGB and depth data recorded with a Microsoft Kinect camera, inertial data from a wearable inertial sensor and skeleton data. The latter was extracted using the Microsoft Kinect v2 API and included a total a 25 joints and their positions in world coordinates (x, y and z). A total of 8 subjects (4 males and 4 females) performed 27 activities, %
each of which was repeated 4 times, producing a total of 861 samples. {Berkeley MHAD} is also aimed at furthering the research in algorithms for recognising human movements. The modalities recorded include RGB and depth using a Microsoft Kinect camera, accelerometers, skeleton data and audio. Skeleton data for this dataset was obtained using the motion capture system Impulse (PhaseSpace Inc., San Leandro, CA), which recorded 3D positions of 48 LED markers that were arranged on the body of the participants. %
The dataset includes 12 subjects performing 11 actions, each repeated 5 different times, for a total of 660 action sequences. {Both datasets include activities like walk, sit-to-stand and a variety of arm movements, which are all typical of activities normally monitored in AAL applications.}

\subsection{Implementation details}
As already mentioned earlier, although our framework {can easily be extended and applied to a variety of different sensor modalities,} %
in this work we focus only  %
on inertial data supplemented with RGB videos and skeletons. For the inertial data, after converting the accelerations into images, as explained in Section~\ref{sec:preprocessing}, we use Inception V3 \cite{Szegedy2016} pre-trained on ImageNet as a backbone. For videos, we use I3D Inception \cite{Carreira2017} as a backbone, with weights pre-trained on Kinetics \cite{kay2017kinetics} and inflated kernels of ImageNet \cite{deng2009imagenet}. The entire method was implemented using Tensorflow and Keras in Python 3.7 and trained using 4 Nvidia GeForce RTX 2070 GPUs\footnote{Our code will be made available at \url{https://github.com/ale152/inertial-lupi}}.

\section{Experiments and results}
\label{sec:results}
We perform the majority of our {experiments and ablations} on the UTD-MHAD dataset and use Berkeley MAHD as a benchmark for further comparisons with other methodologies. {First, we investigate the classification performances of each data modality -- inertial, RGB and skeleton -- when used solo in a single-stream network architecture. Then, we  study the performances of different fusion strategies, using different combinations of these modalities. Finally, we replace the video modality from the fused stream with its hallucinated counterpart and check which hallucination mode produces the least drop in performance.}%

\subsection{Single stream}
Here, we explore the classification performance of each modality when used individually in a single-stream network configuration (see  Table~\ref{tab:results_fusion2}). {As already shown in previous works \cite{Tao2015}, visual sensors outperform inertial-based classifiers, which justifies the wide-spread use of cameras for human motion analysis.} Interestingly, the {visual} modality performing best in a single-stream architecture is the skeleton, yielding an accuracy of 93.16\%, {with colour only being slightly lower at 92.93\%}. This {difference} can be attributed to the nature of the actions from the UTD-MHAD dataset, which as activities of daily living, are mainly accentuated by the movement of the subjects, rather than the background or the objects they interact with\footnote{Different type of activities like sports or outdoor actions might be better characterised by visual elements like background and objects around the subjects.}. Moreover, the activities performed in UTD-MHAD dataset are acted {and no objects  are interacted with},  reducing the helpfulness of visual cues that might be provided by the RGB stream. On the other hand, the inertial sensors alone may struggle to discriminate between actions that are quite similar without pose information, like ``Sit to stand'', ``Squat'' and ``Pickup and throw''.

\begin{table*}[h]
 \caption{%
 Accuracy measures (\%) reported for different modalities adopted individually or in a fusion strategy. Results are reported for the UTD-MHAD dataset using the original train/test split (subjects 1, 3, 5, 7 used for training; subjects 2, 4, 6, 8) used for testing.}
 \centering
\begin{tabular}{l|c|c|c|}
\hline
\multicolumn{1}{|c|}{\textbf{Modality(s)}}         & \multicolumn{3}{c|}{\textbf{Accuracy}}        \\ \hline \hline
\multicolumn{1}{|l|}{Inertial}                     & \multicolumn{3}{c|}{83.49\%}                  \\ \hline
\multicolumn{1}{|l|}{Colour}                       & \multicolumn{3}{c|}{92.93\%}                  \\ \hline
\multicolumn{1}{|l|}{Skeleton}                     & \multicolumn{3}{c|}{93.16\%}                  \\ \hline \hline
\multicolumn{1}{|l|}{\textbf{Fusion Strategy}}   & 
                {\bf Late}      & {\bf Mid Concatenate}  & {\bf Mid Dense}  \\ \hline
\multicolumn{1}{|l|}{Inertial + Skeleton}          & 94.34\%   & 95.76\%            & 95.76\%      \\ \hline
\multicolumn{1}{|l|}{Inertial + Skeleton + Colour} & 96.70\%   & 97.17\%            & \textbf{97.41}\%      \\ \hline
\end{tabular}
  \label{tab:results_fusion2}
\end{table*}

\subsection{Fused stream}
For the fused architecture, we tested the three different strategies described in Section~\ref{sec:fusion_strategies} using two combinations of modalities: inertial+skeleton+colour and a lighter version comprising only inertial+skeleton. 
For the combination of the three streams, the simplest Late fusion strategy produced a combined accuracy of 96.70\%, which is 3.54\% better than the best single modality model. However, the more advanced strategies, Mid Dense and Mid Concatenate, outperformed Late fusion at 97.41\% and 97.17\% respectively. Reducing the number of streams to only two, Inertial and Skeleton, reduces the overall best performance to 95.76\%. Although the performance of the two-streams network is lower than the three-streams one, processing videos is very expensive computationally and the user might accept a small drop in performance as a compromise for the lower computational time and cost.

\begin{table*}[h]
 \caption{%
 Accuracy measures (\%) for different combinations of hallucination modes and fusion strategies, using only the skeleton as a privileged modality. Results are reported using only inertial data for testing on the UTD-MHAD dataset.}%
 
 \centering
\begin{tabular}{r|c|c|c|c|}
\cline{2-5} 
\multicolumn{1}{l|}{\textit{Two-stream model}}   & \multicolumn{2}{c|}{Regression}                           & \multicolumn{2}{c|}{Triplet}                               \\ \cline{2-5} 
\multicolumn{1}{l|}{}                   & \multicolumn{2}{c|}{\textbf{Hallucination mode}}                   & \multicolumn{2}{c|}{\textbf{Hallucination mode}}  \\ \hline %
\multicolumn{1}{|r|}{\textbf{Fusion strategy}}   & Individual                  & Integrated                      & Individual                  & Integrated                       \\ \hline
\multicolumn{1}{|r|}{Baseline inertial} & \multicolumn{4}{c|}{83.49\%}     \\ \hline
\multicolumn{1}{|r|}{Late}              & 87.03\% & N/A                         & 89.39\% & N/A                          \\ \hline
\multicolumn{1}{|r|}{Mid Concatenate} & 88.21\% & N/A                         & \textbf{90.09}\% & N/A                 \\ \hline
\multicolumn{1}{|r|}{Mid Dense}       & 87.97\% & 86.79\% & 88.92\% & 85.38\% \\ \hline
\end{tabular}
  \label{tab:results_hallucinated_2s}
\end{table*}

\subsection{Hallucinated modes}
In this experiment, we analyse \textit{Individual} and \textit{Integrated} hallucination modes when combined with different fusion techniques as listed earlier in Table~\ref{tab:fusion_hall_modes}. For each combination, we compare our novel triplet-based hallucination learning with a classical regression strategy as per \cite{Hoffman2016}. The learning stage involves both the inertial and skeleton streams and the inference time results, using solely the inertial stream, are shown in Table~\ref{tab:results_hallucinated_2s}. Our triplet learning strategy supercedes regression learning for all fusion strategies of the individually hallucinated features. When compared to the baseline inertial accuracy of 83.49\%, our method achieves an improvement of 6.60\%.

Our integrated hallucination model achieves 86.79\% when trained by regression and 85.38\% by triplets. While these are just competitively below the individual hallucination outcomes, they do come at lower computational cost and training time, and can be a suitable solution for AAL applications.

Similarly, we present our results for the three-stream training process, followed by inference by inertial data only, in Table~\ref{tab:results_hallucinated_3s}.  Our triplet hallucination strategy again achieves the best performances with the same highest level of accuracy of 90.09\%, but this time for the Late fusion strategy. The integrated hallucination mode improved from 85.38\% in the two-stream training model to 87.74\% in this model. This validates the use of an integrated hallucination model, which comes at a relatively lower cost, especially when three sensor types are deployed.

\begin{table*}[h]
 \caption{Accuracy measures (\%) for different combinations of hallucination modes and fusion strategies, using both RGB and skeleton modalities as privileged information. Results are reported using only inertial data for testing on the UTD-MHAD dataset.} 
  \centering
\begin{tabular}{r|c|c|c|c|}
\cline{2-5}
\multicolumn{1}{l|}{}                          & \multicolumn{4}{c|}{\textbf{Learning Mode}}                                                         \\ \cline{2-5} 
\multicolumn{1}{l|}{\textit{Three-stream model}}        & \multicolumn{2}{c|}{Regression}                  & \multicolumn{2}{c|}{Triplet}                  \\ \cline{2-5} 
\multicolumn{1}{l|}{}                          & \multicolumn{2}{c|}{\textbf{Hallucination mode}} & \multicolumn{2}{c|}{\textbf{Hallucination mode}} \\ \hline \hline
\multicolumn{1}{|r|}{\textbf{Fusion strategy}} & Individual                & Integrated       & Individual     & Integrated               \\ \hline
\multicolumn{1}{|r|}{Baseline inertial} & \multicolumn{4}{c|}{83.49\%}  \\ \hline   
\multicolumn{1}{|r|}{Late}                     & 89.39\%                    & N/A                  & \textbf{90.09}\% & N/A    \\ \hline
\multicolumn{1}{|r|}{Mid Concatenate}        & 85.85\%                    & N/A                  & 88.92\%                    & N/A                  \\ \hline
\multicolumn{1}{|r|}{Mid Dense}              & 87.03\%                    & 87.50\% & 88.44\% & 87.74\% \\ \hline
\end{tabular}
  \label{tab:results_hallucinated_3s}
\end{table*}

\subsection{Inference classes different from training classes}
In this experiment, we assess the generalisation of the features learnt through the hallucination process {by using a different subset of action classes during training  and testing.} %
{First, we divided the UTD-MHAD dataset in two subsets by arbitrarily keeping the first half of action classes 1 to 14 and discarding the remaining 15 to 27 labels}\footnote{Full description of the action labels can be found at \cite{Chen2015a}}. %
{Using the inertial and skeleton modalities from this reduced dataset, we trained single stream, fusion and hallucination models as per previous Sections. Rather than testing all the models again, we selected the best combination of Individual hallucination mode with Mid Concatenate fusion strategy from Table~\ref{tab:results_hallucinated_2s}}. These models were able to achieve 86.16\% accuracy using the inertial stream, 95.54\% using the skeletons and 97.32\% through the fusion strategy, with results similar to those seen in Table~\ref{tab:results_fusion2} for the full set of classes. After this stage, the hallucination model is frozen.

In a second stage, we once again divided the UTD-MHAD dataset using action labels, but this time keeping the second half of the action classes 15 to 27, while discarding the first half. We then calculated the hallucinated features from the hallucination model trained during the first stage, and the classifier was able to produce an accuracy of 92.00\%. In comparison, a standard inertial classifier trained on the same set of classes 15 to 27, only produced an accuracy of 90.50\%, which is a reduction of 1.5\%. %
This improvement is lower than the one experienced in the previous experiment, probably also due to the size of the training dataset that is halved when selecting only a subset of the actions, but it nonetheless confirms the validity and generalisation power of our method.

\subsection{Cross-validation}

{We also assessed our method through a cross-validation strategy, where we trained the single streams, fusion and hallucination models eight independent times, leaving one subject out for validation each time. The results in Table~\ref{tab:cross-validation} confirm once again that the hallucinated model, at an average accuracy of 95.07\% performs better than when using an inertial sensor alone and can be a significantly close approach in comparison to using more sensors.}

\begin{table*}[h]
 \caption{Accuracy measures (\%) for each tested modality for all the individual subjects of the UTD-MHAD dataset.}
  \centering
\begin{tabular}{c|c|l|l|c|}
\cline{2-5}
\multicolumn{1}{l|}{}                  & \multicolumn{4}{c|}{\textbf{Modality}}                                          \\ \hline
\multicolumn{1}{|c|}{\textbf{Subject}} & \textbf{Inertial} & \textbf{Skeleton} & \textbf{Fusion} & \textbf{Hallucinated} \\ \hline \hline
\multicolumn{1}{|c|}{1}            & 92.31\%           & 100.00\%          & 100.00\%        & 96.15\%               \\ \hline
\multicolumn{1}{|c|}{2}            & 92.31\%           & 100.00\%          & 100.00\%        & 97.17\%               \\ \hline
\multicolumn{1}{|c|}{3}            & 87.50\%           & 100.00\%          & 100.00\%        & 94.23\%               \\ \hline
\multicolumn{1}{|c|}{4}            & 91.35\%           & 97.12\%           & 98.08\%         & 95.19\%               \\ \hline
\multicolumn{1}{|c|}{5}            & 90.38\%           & 99.04\%           & 100.00\%        & 95.19\%               \\ \hline
\multicolumn{1}{|c|}{6}            & 92.31\%           & 100.00\%          & 100.00\%        & 95.19\%               \\ \hline
\multicolumn{1}{|c|}{7}            & 91.35\%           & 96.15\%           & 97.12\%         & 95.19\%               \\ \hline
\multicolumn{1}{|c|}{8}            & 88.46\%           & 98.08\%           & 99.04\%         & 92.31\%               \\ \hline \hline
\multicolumn{1}{|c|}{\bf Average}          & 90.75\%           & 98.80\%           & 99.28\%         & 95.07\%               \\ \hline
\end{tabular}
  \label{tab:cross-validation}
\end{table*}

\subsection{Comparison with state of the art}
{Given our results are obtained using inertial data only, we  compare against other methods that employ inertial sensors during inference. The results in Table~\ref{tab:comparison_sota} again substantiate the superiority of our proposed approach (for both different validation protocols) against existing state-of-the-art approaches on UTD-MHAD, where we reach an accuracy of 95.07\% on the cross-validation protocol and 90.09\% on the original split.}

\begin{table*}[h]
 \caption{Comparative results of our hallucinated inertial approach against other works that apply the inertial modality only on the UTD-MHAD dataset -- for both cross-validation and the original training/testing split from \cite{Chen2015a} protocols.}
  \centering
\begin{tabular}{|l|c|lc}
\hline
\multicolumn{2}{|c|}{\bf Cross-validation}                                   & \multicolumn{2}{c|}{\bf Original split (train: 1,3,5,7, test: 2,4,6,8)}                         \\ \hline
\multicolumn{1}{|c|}{\textbf{Method}}                & \textbf{Accuracy} & \multicolumn{1}{c|}{\textbf{Method}}               & \multicolumn{1}{c|}{\textbf{Accuracy}} \\ \hline \hline
Imran et al. \cite{Imran2020}                        & 86.51\%           & \multicolumn{1}{l|}{Moddrop \cite{Neverova2016}}   & \multicolumn{1}{c|}{83.73\%}          \\ \hline
Faud et al. \cite{Fuad2018}                          & 81.2\%            & \multicolumn{1}{l|}{Chen et al. \cite{Chen2015a}}  & \multicolumn{1}{c|}{67.2\%}            \\ \hline
Chen et al. \cite{Chen2016c}                         & 76.4\%            & \multicolumn{1}{l|}{Dawar et al. \cite{Dawar2017}} & \multicolumn{1}{c|}{86.3\%}            \\ \hline
Wei et al. \cite{Wei2019}                            & 90.3\%            & \multicolumn{1}{l|}{Baseline inertial (Ours)}      & \multicolumn{1}{c|}{83.49\%}           \\ \hline
Aparecida et al \cite{Aparecida2021}                 & 82.23\%           & \multicolumn{1}{l|}{Hallucinated (Ours)}           & \multicolumn{1}{c|}{\textbf{90.09\%}}   \\ \hline
Ahmad et al. \cite{Ahmad2019}                        & 93.7\%\footnote{Random split}&                                                    &                                        \\ \cline{1-2}
Baseline inertial (Ours)                             & 90.75\%           &                                                    &                                        \\ \cline{1-2}
Hallucinated (Ours)                                  & \textbf{95.07}\%  &                                                    &                                        \\ \cline{1-2}
\end{tabular}
  \label{tab:comparison_sota}
\end{table*}

{We performed a similar comparative evaluation on the Berkeley MHAD dataset with the results shown in Table~\ref{tab:results_berkeley_sota}. We can reach similar conclusions as before on the improvements made on the state of the art. For example, for the cross-validation protocol, our hallucinated approach reached 99.31\% accuracy, and when training on subjects 1 to 7 and testing on subjects 8 to 12 as in the original split from \cite{ofli2013berkeley}), it achieved 91.54\%.}

\begin{table*}[h]
 \caption{Comparative results of our hallucinated inertial approach against other works that apply the inertial modality only on the UTD-MHAD dataset -- for both cross-validation and the original training/testing split protocols as per \cite{Chen2015a}.
 }
  \centering
\begin{tabular}{lc|l|c|}
\hline
\multicolumn{4}{|c|}{\textbf{Validation protocol}}                                                                          \\ \hline
\multicolumn{2}{|c|}{Cross-validation}                       & \multicolumn{2}{c|}{Original split (train: 1-7, test: 8-12)} \\ \hline
\multicolumn{1}{|c|}{\textbf{Method}}                      & \textbf{Accuracy} & \multicolumn{1}{c|}{\textbf{Method}}             & \textbf{Accuracy}           \\ \hline
\multicolumn{1}{|l|}{Chen et al. \cite{Chen2015}} & 92.01\%  & Moddrop \cite{Neverova2016}                    & 87.87\%               \\ \hline
\multicolumn{1}{|l|}{Baseline inertial (Ours)}    & 98.70\%  & Ofli et al. \cite{ofli2013berkeley}            & 85.40\%            \\ \hline
\multicolumn{1}{|l|}{Hallucinated (Ours)}         & \textbf{99.31\%}  & Baseline inertial (Ours)                & 86.03\%            \\ \hline
                                                  &          & Hallucinated (Ours)                     & \textbf{91.54\%}            \\ \cline{3-4} 
\end{tabular}
  \label{tab:results_berkeley_sota}
\end{table*}

\subsection{Limitations}
The experiments suggest that our proposed methodology is very effective and, indeed, accelerometer data can be supplemented with RGB or skeleton data to improve classification accuracy for activities of daily living. Although this methodology can be extended to the classification of a different set of labels (and potentially different modalities), an inherent requirement for our method to work is the overlap of information across different modalities. For example, if for a hand movement, the information provided by the accelerometer sensor is present in the RGB data and in the form of joint position in the skeleton data, then it is possible to leverage the latter two to improve the classification of the former. Were modalities to be completely independent and uncorrelated, we would expect the algorithm not to work as well as in our circumstances.

\subsection{Computational effort}
Computational efficiency is a very important aspect in our methodology. Algorithms for AAL often need to run on small, cheap hardware that is deployable in people's homes, where computation resources are constrained and internet connectivity might be limited, hindering access to cloud resources. These reasons underlay our development of the very efficient Integrated Modality Hallucination approach presented in Section~\ref{sec:modality_hallucination}. 

Like with any other deep learning model, the training stage is comparatively the most computationally expensive part of the process. In our case, we used an Ubuntu system with 4 NVIDIA GeForce RTX 3070 GPUs running Python and Tensorflow. The  training of the video classifier  took up to a week, and  the skeleton and inertial classifiers took a day each. Once the classifiers were optimised, training of the hallucination networks was very efficient, taking less than 12 hours, thanks to the pre-computation of the features described in Section~\ref{sec:training_hall_net}. %

\begin{table*}
 \caption{Inference time in ms per clip.}
  \centering
\begin{tabular}{@{}rc@{}}
\toprule
\multicolumn{1}{c}{\textbf{Method}} & \textbf{Time (ms/clip)} \\ \midrule
Inertial (I)                        & 25.4                   \\
Colour (C)                          & 129                    \\
Skeleton (S)                        & 24.3                   \\
Late Fusion (ICS)                   & 167                    \\
Mid Concatenate (ICS)               & 169                    \\
Mid Dense (ICS)                     & 173                    \\
Individual Hallucination (I)        & 76.2                   \\
Integrated Hallucination (I)        & 25.4                   \\ \bottomrule
\end{tabular}
  \label{tab:computational_effort}
\end{table*}

The time required to classify a clip at the inference stage is reported in Table~\ref{tab:computational_effort} for each individual modality, different fusion strategies and our hallucinated proposals. Inertial and Skeleton modalities are  faster than Colour by a factor of 5, which is to be expected due to the heavy architecture used to classify videos. The three fusion strategies are very comparable with insignificant differences. The Hallucination strategies do not need to process a computationally expensive video stream and our network architecture  for these models is identical to the standard acceleration classifier, hence the inference time to run our Integrated Modality Hallucination is identical to the  accelerometer-only classifier. This make our Hallucination model not only an efficient alternative to the problem of missing modality, but also a potential replacement of modality altogether, in case of strict performance requirements.

\section{Conclusions}
The vast majority of AAL technologies involve a variety of sensor modalities that work together to monitor and support people's health in their own homes. {Although fusing different modalities improves the overall performances of monitoring algorithms, standard fusing approaches reduce the overall reliability of multi-sensory frameworks. Moreover,} each sensing device has a limited field of view and multi-sensory algorithms can only work at the intersection of those, strongly limiting their possibilities. {In this paper, we propose a novel solution to sensory fusion for AAL that takes advantage of the concepts of privileged information and modality hallucination. The main limitations of standard fusion approaches arise from the mandatory availability, at inference time, of all sensing modalities employed during training. Thanks to modality hallucination, we can train an algorithm to use all the available modalities during training, and then replace missing modalities with an ``hallucinated'' version during inference. }%
{For the first time, we propose an AAL framework %
that involves inertial data from a wearable device supplemented with privileged information from the vision monitoring system, namely RGB videos and skeletons. We propose a novel strategy for modality hallucination that combines the classical triplet learning with modality hallucination, showing successful results in an AAL scenario.} In particular, we show that an activity classifier trained with our framework and using inertial data only during inference is able to improve its accuracy by 7 percentage points on the UTD-MHAD dataset and 6 percentage points on the Berkeley MHAD dataset, compared to a similar algorithm that is trained in a standard framework. We also show state-of-the-art results for action classification using only inertial data during inference for both datasets.

\section*{Acknowledgments}
This work was performed under the SPHERE Next Steps Project funded by the UK Engineering and Physical Sciences Research Council (EPSRC), Grant EP/R005273/1.

\bibliographystyle{unsrtnat}
\bibliography{references}  %

\end{document}